\documentclass{article}

\usepackage{template/spconf,amsmath,amssymb,graphicx,hyperref}
\usepackage{booktabs}
\usepackage{microtype}
\usepackage{multirow}
\usepackage{color}

\title{Dual-Scale State-Space Modeling with Speaker-Wise Dynamic CRF for Conversational Speech Emotion Recognition}

\name{Guan-Hua Wen  \qquad Hou-Chiang Tseng  \qquad Kuan-Yu Chen }
\address{National Taiwan University of Science and Technology, Taipei, Taiwan}

\begin{document}
\ninept

\maketitle

\begin{abstract}
Conversational speech emotion recognition requires both dialogue-wide contextual modeling and speaker-specific emotion tracking. We propose DSSM-CRF, an audio-only framework that separates the structural roles of these two dependencies. Self-supervised speech representations are processed by bidirectional state-space encoders at frame and dialogue scales to capture local acoustic dynamics and conversational context. The contextualized utterances are then organized into separate speaker-wise chains for structured decoding. A dynamic conditional random field combines a global transition matrix with pair-conditioned transition residuals, while an auxiliary emotion-shift objective supervises the shared edge representation during training. This design allows interlocutor turns to influence contextual representations without treating them as transitions in another speaker's emotion trajectory. DSSM-CRF achieves 75.81\% UA and 74.90\% WA on IEMOCAP, and 54.72\% WA and 49.31\% WF1 on MELD. Ablations and matched controls support the contributions of speaker-wise factorization, contextual transition modeling, and structured decoding.
\end{abstract}


\section{Introduction}
\label{sec:introduction}

Speech emotion recognition (SER) aims to infer affective states from speech and supports applications such as affect-aware assistants, call-center analytics, and behavioral monitoring. Earlier SER systems relied on hand-crafted acoustic descriptors and conventional classifiers~\cite{elayadi2011survey}, whereas recent self-supervised speech models provide transferable representations directly from raw waveforms~\cite{pepino2021wav2vecser}. Conversational SER, however, differs from isolated-utterance classification because the emotion expressed in a turn depends not only on its acoustic realization but also on surrounding interactions and the speaker's emotional history. An effective audio-only system must therefore model local acoustic dynamics, dialogue context, and speaker-specific emotion evolution.

A central challenge is that dialogue context and emotion evolution play different structural roles: an interlocutor's response can inform the current emotion but is not part of that speaker's trajectory. Existing recurrent speaker-state models~\cite{majumder2019dialoguernn}, interaction graphs~\cite{ghosal2019dialoguegcn}, and dialogue-aware pretrained models~\cite{shen2021dialogxl} capture intra- and inter-speaker dependencies~\cite{poria2019ercsurvey}, while CRF-based methods model dependencies among emotion labels~\cite{lafferty2001crf}, including emotion shifts~\cite{esa_crf}. However, a single dialogue-level CRF chain conflates cross-speaker responses with within-speaker transitions, and a shared transition matrix cannot adapt to individual utterance pairs. SCoPE's speaker-conditioned priors and shift-aware fusion further highlight the importance of speaker-specific emotion history~\cite{SCoPE2026}.

We therefore use all turns for dialogue-wide contextualization but restrict structured transitions to each speaker's own sequence. These dependencies also span different temporal resolutions, from short prosodic patterns within utterances to long conversational context. Hierarchical speech models have explored multi-scale temporal modeling~\cite{dst,dwformer2023,speechformer_pp}, while structured and selective state-space models provide efficient sequence modeling~\cite{gu2022s4,gu2024mamba}.

Based on this formulation, we propose \textbf{DSSM-CRF}, an audio-only framework that applies bidirectional state-space encoders to WavLM representations~\cite{chen2022wavlm} at frame and dialogue scales before forming same-speaker CRF chains. Each edge combines a global transition matrix with a pair-conditioned residual, while an auxiliary emotion-shift objective supervises the shared edge representation.

Our main contributions are:
\begin{itemize}
\item We formulate conversational SER by separating dialogue-wide contextual influence from speaker-specific emotion evolution: all turns contribute to contextual representations, while structured transitions follow each speaker's own trajectory.
\item We propose a speaker-wise dynamic CRF that combines a global transition matrix with pair-conditioned residual transitions and auxiliary emotion-shift supervision.
\item We develop dual-scale bidirectional state-space modeling for frame-level acoustic dynamics and dialogue-level context, and validate the framework on IEMOCAP and MELD through component ablations, matched structural controls, and backbone analyses.
\end{itemize}

\begin{figure}[t]
	\centering
    \includegraphics[width=\columnwidth]{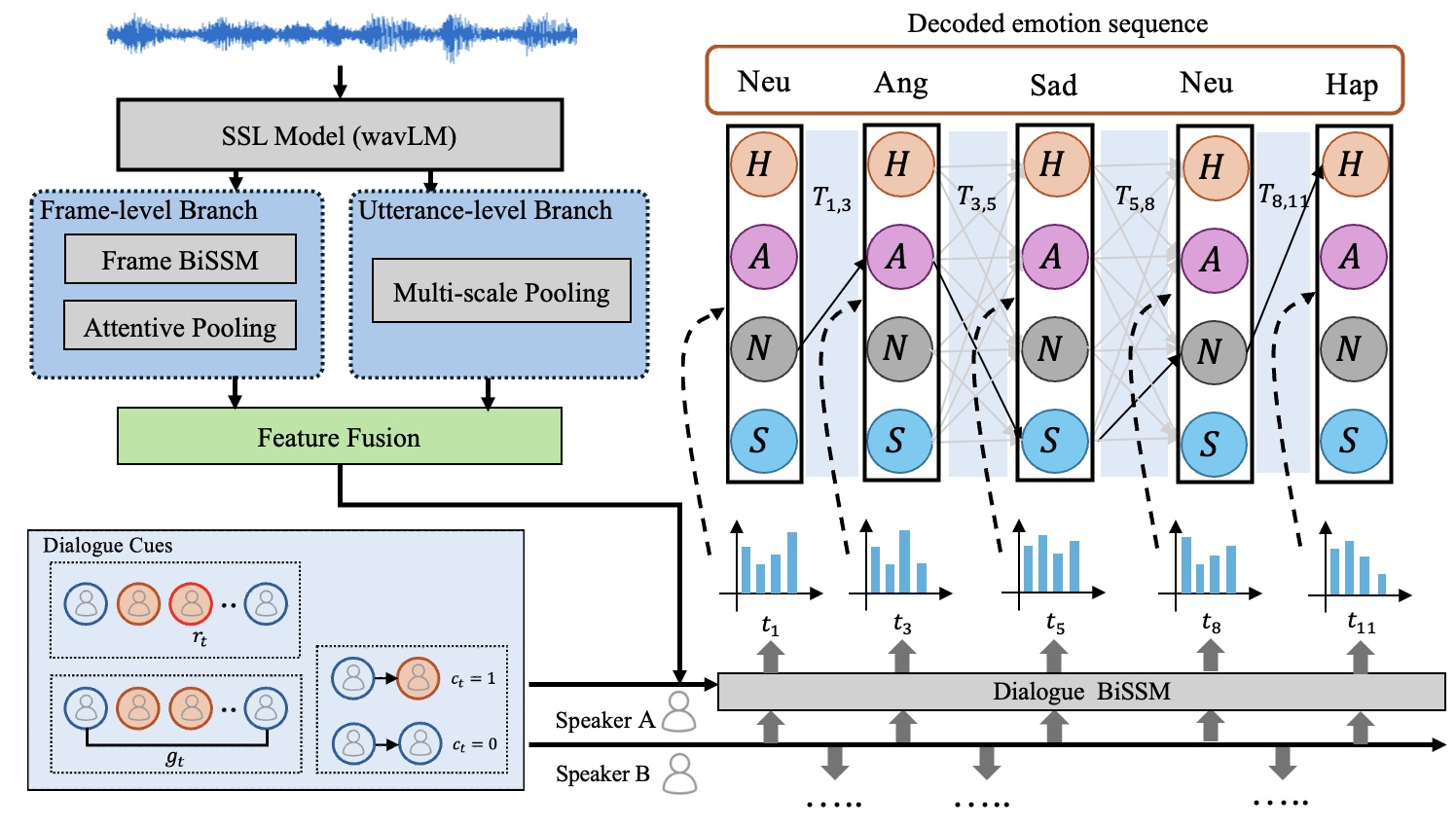}
	\caption{Overview of DSSM-CRF. Frame- and utterance-level representations are fused, augmented with dialogue cues, and contextualized by the dialogue BiSSM. Speaker-wise CRF decoding uses emotion emissions and pair-conditioned transitions along each speaker's chain.}
	\label{fig:architecture}
\end{figure}

\section{Methodology}
\label{sec:method}

\subsection{Problem formulation}

Let a dialogue be $\mathcal{U}=\{(x_t,s_t,y_t)\}_{t=1}^{T}$, where $x_t$ is the waveform of turn $t$, $s_t$ denotes its speaker, and $y_t\in\{1,\ldots,K\}$ is its emotion label. Given the waveforms and speaker identities of all turns, the goal is to infer $\{y_t\}_{t=1}^{T}$ while exploiting dialogue-wide contextual information and modeling emotion transitions along each speaker's own trajectory.

\subsection{Acoustic representation}

For utterance $t$ with $F_t$ frames, WavLM-Large~\cite{chen2022wavlm}
produces hidden sequences. Denoting its final twelve hidden layers as
$H_t^{(\ell)}\in\mathbb{R}^{F_t\times D}$, $\ell=1,\ldots,12$,
DSSM-CRF learns their convex combination:
\begin{equation}
H_t=\sum_{\ell=1}^{12}\omega_\ell H_t^{(\ell)}, \quad
\omega_\ell=\frac{\exp(\alpha_\ell)}
{\sum_{j=1}^{12}\exp(\alpha_j)},
\label{eq:layer_fusion}
\end{equation}
where $\alpha_\ell$ is a learned scalar and $H_t\in\mathbb{R}^{F_t\times D}$ is the fused frame sequence.

To preserve coarse temporal structure within an utterance, multi-scale attentive statistics pooling divides the frame sequence into one, two, and four contiguous temporal partitions. For each partition, attention-weighted mean and standard deviation are computed. The seven descriptors from the $1+2+4$ partitions are concatenated and projected to obtain an utterance representation $e_t\in\mathbb{R}^{256}$.

\subsection{Dual-scale state-space modeling}

At the frame scale, $H_t$ is projected to the common model dimension $d=256$ and processed by a bidirectional Mamba-2-style state-space encoder~\cite{dao2024mamba2}. Independent forward and backward stacks process the sequence in opposite directions, and the backward output is restored to chronological order. Attentive pooling of the contextualized frame sequence produces $f_t$, which is fused with $e_t$:
\begin{equation}
h_t=\operatorname{LN}\!\left(
\operatorname{Drop}\!\left(
\operatorname{GELU}\!\left(W_f[e_t;f_t]+b_f\right)
\right)\right).
\label{eq:fusion}
\end{equation}

At the dialogue scale, each utterance representation is augmented with three structural cues. The normalized dialogue position is $r_t=(t-1)/\max(T-1,1)$. Let $p_t=\max\{k<t:s_k=s_t\}$ denote the preceding dialogue position spoken by the same speaker when it exists. The same-speaker turn gap is $g_t=t-p_t$, with $g_t=0$ for the speaker's first turn, and is normalized as $\tilde{g}_t=\log(1+g_t)/\log(2+T)$. We further define the speaker-change indicator as $c_1=0$ and $c_t=\mathbb{1}[s_t\neq s_{t-1}]$ for $t>1$. The dialogue input is
\begin{equation}
q_t=h_t+W_r r_t+W_g\tilde{g}_t+E_{c_t},
\label{eq:dialogue_input}
\end{equation}
where $W_r,W_g\in\mathbb{R}^{d\times 1}$ are learned projections and $E_{c_t}\in\mathbb{R}^{d}$ is a learned speaker-change embedding.

Given the ordered dialogue sequence $\mathbf{q}=(q_1,\ldots,q_T)$, the dialogue-level contextual representation is
\begin{equation}
z_t=\operatorname{LN}\!\left(
h_t+\operatorname{BiSSM}_{\mathrm{dlg}}(\mathbf{q})_t
\right).
\label{eq:dialogue}
\end{equation}
Both the frame- and dialogue-level encoders contain independent two-layer forward and backward state-space stacks. The frame recurrence is reset for each utterance, whereas the dialogue recurrence spans the complete ordered conversation. The outputs of the forward and backward stacks are concatenated after restoring the backward sequence to chronological order, and are mapped from $2d$ to $d$ by a linear--GELU--dropout--layer-normalization fusion. Consequently, each $z_t$ incorporates contextual information from both the current speaker and interlocutors before structured decoding.

\subsection{Speaker-wise dynamic CRF}

For each speaker $s$, let $\pi_s=(t_1,\ldots,t_{M_s})$ denote the ordered dialogue positions of that speaker's utterances. Each $\pi_s$ defines a separate CRF chain for structured decoding. The emotion emission scores at turn $t$ are
\begin{equation}
\boldsymbol{\eta}_t=W_e z_t+b_e\in\mathbb{R}^{K}.
\label{eq:emission}
\end{equation}

For each consecutive same-speaker pair $(i,j)$, we construct an edge representation as
\begin{equation}
u_{i,j}=\phi\!\left([z_i;z_j;|z_j-z_i|;z_i\odot z_j]\right),
\label{eq:edge}
\end{equation}
where $\phi:\mathbb{R}^{4d}\rightarrow\mathbb{R}^{d}$ is a linear--GELU--dropout projection. The edge representation predicts both a pair-conditioned transition residual and an emotion-shift logit:
\begin{equation}
\begin{split}
\Delta T_{i,j}
&=\operatorname{reshape}\!\bigl(W_Tu_{i,j}+b_T\bigr), \\
T_{i,j}&=T_{\mathrm{global}}+\Delta T_{i,j}, \\
\rho_{i,j}&=w_{\mathrm{shift}}^{\top}u_{i,j}+b_{\mathrm{shift}},
\end{split}
\label{eq:transition}
\end{equation}
where $W_T\in\mathbb{R}^{K^2\times d}$, $b_T\in\mathbb{R}^{K^2}$, and $w_{\mathrm{shift}}\in\mathbb{R}^{d}$. The flattened transition residual is reshaped as $\Delta T_{i,j}\in\mathbb{R}^{K\times K}$. Here $T_{\mathrm{global}}\in\mathbb{R}^{K\times K}$ is a learned transition matrix shared across all same-speaker edges, while $\Delta T_{i,j}$ provides pair-dependent adjustments based on the contextualized utterances.

For a label sequence $\mathbf{y}_s$ along speaker chain $\pi_s$, its score is
\begin{equation}
\begin{split}
\operatorname{Score}_s(\mathbf{y}_s)
&=\sum_{m=1}^{M_s}
\boldsymbol{\eta}_{t_m}[y_{t_m}] \\
&\quad+
\sum_{m=2}^{M_s}
T_{t_{m-1},t_m}
[y_{t_{m-1}},y_{t_m}].
\end{split}
\label{eq:crf_score}
\end{equation}
The normalized CRF negative log-likelihood is
\begin{equation}
\begin{split}
\mathcal{L}_{\mathrm{CRF}}
&=\frac{1}{|\mathcal{S}|}
\sum_{s\in\mathcal{S}}
\frac{1}{M_s}
\Bigg(
\log\sum_{\mathbf{y}'_s}
\exp\!\left(\operatorname{Score}_s(\mathbf{y}'_s)\right) \\
&\qquad\qquad
-\operatorname{Score}_s(\mathbf{y}_s)
\Bigg),
\end{split}
\label{eq:crf_nll}
\end{equation}
where $\mathcal{S}$ denotes the set of speakers. For $M_s=1$, the transition term is empty and decoding depends only on the emission score. Viterbi decoding is performed independently for each speaker chain.

This factorization preserves interlocutor information because each $z_t$ is contextualized using the complete dialogue. It restricts only the structured transition edges to a speaker's own utterance sequence, while cross-speaker interactions can still affect both emotion emissions and pair-conditioned transition scores through the contextualized representations.

\subsection{Training objective}

For each same-speaker edge $(i,j)$, the shift target is derived only from training labels as $b_{i,j}=\mathbb{1}[y_i\neq y_j]$. For a dialogue with $E>0$ same-speaker edges, including $E_+$ shift cases and $E_-$ inertia cases, we use the balanced auxiliary loss
\begin{equation}
\begin{split}
\mathcal{L}_{\mathrm{shift}}
&=-\frac{1}{E}\sum_{(i,j)} w_{b_{i,j}}
\Big[
b_{i,j}\log\sigma(\rho_{i,j}) \\
&\quad+
(1-b_{i,j})
\log\!\left(1-\sigma(\rho_{i,j})\right)
\Big].
\end{split}
\label{eq:shift_loss}
\end{equation}
When $E_+>0$ and $E_->0$, $w_1=E/(2E_+)$ and $w_0=E/(2E_-)$. We set $\mathcal{L}_{\mathrm{shift}}=0$ when $E=0$ and use unweighted binary cross entropy when only one edge type is present.

The overall training objective is
\begin{equation}
\mathcal{L}
=\mathcal{L}_{\mathrm{CRF}}
+\lambda_{\mathrm{shift}}\mathcal{L}_{\mathrm{shift}}.
\label{eq:objective}
\end{equation}
The shift objective supervises the edge representation during training but does not participate in Viterbi inference.

\section{Experimental Setup}
\label{sec:experiments}

\subsection{Datasets and evaluation}

IEMOCAP~\cite{busso2008iemocap} is evaluated in the four-class setting with angry, happy (including excited), sad, and neutral emotions. We use five-fold session rotation with three sessions for training, one for validation, and one for testing. Utterances are ordered by timestamp within each recording, and dialogue modeling and CRF edges do not cross recording boundaries. MELD~\cite{poria2019meld} uses its official seven-class split. Only audio and speaker identities are used; transcripts, sentiment annotations, and visual information are excluded.

We report unweighted accuracy (UA), weighted accuracy (WA), and weighted F1-score (WF1). Utterances with a same-speaker predecessor are additionally divided into Shift and Inertia subsets according to whether the gold emotion changes. Model selection uses validation UA on IEMOCAP and WA on MELD.

\subsection{Implementation details}

Training consists of two stages. First, a class-weighted cross-entropy objective adapts the layer fusion, pooling, classifier, and progressively unfrozen speech encoder, using $w_c=(N/(Kn_c))^\gamma$ with $\gamma=1$ on IEMOCAP and validation-selected $\gamma=0.4$ on MELD. The first three epochs train only task-specific modules, followed by five epochs with the upper half of the speech encoder unfrozen; the entire speech encoder is subsequently unfrozen if the validation score improves. This stage runs for at most 30 epochs with patience 6.

The selected speech encoder is then frozen and the BiSSMs and CRF are trained for at most 50 epochs with patience 8. Each forward and backward state-space stack contains two layers with RMS normalization, a depthwise convolution with kernel size 4, state size 16, expansion factor 2, gated output projection, and residual connections. The model dimension and dropout rate are 256 and 0.1, respectively. Across the two training stages, AdamW uses learning rates of $10^{-5}$ for unfrozen speech-backbone parameters and $3\times10^{-4}$ for task-specific modules, with weight decay $10^{-2}$, gradient clipping at 1.0, and a 120-s effective audio batch. Audio inputs are truncated to 30~s; waveform augmentation uses random gain up to 6~dB and 0.2-s time masking.

For ablations, we individually remove the frame-level SSM, dialogue-level SSM, transition residual, or shift supervision. Speaker-wise topology is isolated by comparing the speaker-wise CRF and a matched single-chain CRF with $\lambda_{\mathrm{shift}}=0$ for both, while structured decoding is compared with a retrained no-CRF model using utterance-level cross entropy while retaining the auxiliary shift objective. We evaluate $\lambda_{\mathrm{shift}}\in\{0,0.1,\ldots,0.5\}$ with fixed WavLM-Large representations and use 0.2 for the reported models. Backbone robustness is evaluated with wav2vec~2.0 Large~\cite{baevski2020wav2vec2}, HuBERT-Large~\cite{hsu2021hubert}, and WavLM-Large~\cite{chen2022wavlm}.

\begin{table}[t]
	\centering
	\caption{Performance comparison on IEMOCAP. $\dagger$ denotes the second-best result.}
	\label{tab:iemocap_4class_comparison}
	\begin{tabular*}{\columnwidth}{@{\extracolsep{\fill}}lcc@{}}
		\toprule
		\textbf{Model} & \textbf{UA(\%)} & \textbf{WA(\%)} \\
		\midrule
		ADAN~\cite{adan} & 64.51 & 66.92 \\
		Zou et al.~\cite{zou2022} & 71.05 & 69.80 \\
		DST~\cite{dst} & 73.60 & 71.80 \\
		DWFormer~\cite{dwformer2023} & 73.90 & 72.30 \\
		Mel-MViTv2~\cite{mel_mvitv2} & -- & 64.03 \\
		SCQT-MaxViT~\cite{scqt_maxvit} & -- & 62.49 \\
		SpeechFormer++~\cite{speechformer_pp} & 71.50 & 70.50 \\
		TIM-Net~\cite{timnet} & 72.50 & 71.65 \\
		Vesper~\cite{vesper} & 74.30$^{\dagger}$ & 73.70$^{\dagger}$ \\
		DFSD-EMO~\cite{dfsd_emo} & 72.35 & 72.36 \\
		\midrule
		\textbf{DSSM-CRF (ours)} & $\mathbf{75.81}$ & $\mathbf{74.90}$ \\
		\bottomrule
	\end{tabular*}
\end{table}

\begin{table}[t]
	\centering
	\caption{Performance comparison on MELD. $\dagger$ denotes the second-best result.}
	\label{tab:meld_comparison}
	\begin{tabular*}{\columnwidth}{@{\extracolsep{\fill}}lcc@{}}
		\toprule
		\textbf{Model} & \textbf{WA(\%)} & \textbf{WF1(\%)} \\
		\midrule
		DST~\cite{dst} & -- & 48.80$^{\dagger}$ \\
		DWFormer~\cite{dwformer2023} & -- & 48.50 \\
		SpeechFormer++~\cite{speechformer_pp} & 51.00 & 47.00 \\
		Vesper~\cite{vesper} & 53.50 & 48.00 \\
		DFSD-EMO~\cite{dfsd_emo} & 53.74$^{\dagger}$ & 47.60 \\
		\midrule
		\textbf{DSSM-CRF (ours)} & $\mathbf{54.72}$ & $\mathbf{49.31}$ \\
		\bottomrule
	\end{tabular*}
\end{table}

\begin{table}[t]
	\centering
    \caption{Component ablations and matched controls. S and I denote the Shift and Inertia subsets, respectively; IEMOCAP reports subset UA and MELD reports subset WA. Bold indicates the best result in each column.}
	\label{tab:ablation}

	\setlength{\tabcolsep}{0.8pt}
	\begin{tabular*}{\columnwidth}{@{\extracolsep{\fill}}lcccc@{}}
		\toprule
		\multicolumn{5}{c}{\textbf{(a) IEMOCAP}} \\
		\midrule
		\textbf{Configuration} & \textbf{UA(\%)} & \textbf{WA(\%)} & \textbf{S-UA(\%)} & \textbf{I-UA(\%)} \\
		\midrule
		\textbf{Full DSSM-CRF} & \textbf{75.81} & \textbf{74.90} & 59.17 & 79.06 \\
		$-$ frame-level SSM & 75.67 & 74.37 & 57.53 & \textbf{79.07} \\
		$-$ dialogue-level SSM & 75.31 & 73.88 & 59.06 & 78.19 \\
		$-$ transition residual & 74.09 & 72.79 & 59.74 & 76.62 \\
		$-$ shift supervision & 75.44 & 74.45 & 58.19 & 78.58 \\
		Single-chain CRF ($\lambda_{\mathrm{shift}}=0$) & 74.45 & 73.54 & 59.66 & 77.09 \\
		No-CRF (retrained) & 74.64 & 73.11 & \textbf{61.03} & 77.25 \\
		\bottomrule
	\end{tabular*}

	\par\vspace{3pt}

	\setlength{\tabcolsep}{0.8pt}
	\begin{tabular*}{\columnwidth}{@{\extracolsep{\fill}}lcccc@{}}
		\toprule
		\multicolumn{5}{c}{\textbf{(b) MELD}} \\
		\midrule
		\textbf{Configuration} & \textbf{WA(\%)} & \textbf{WF1(\%)} & \textbf{S-WA(\%)} & \textbf{I-WA(\%)} \\
		\midrule
		\textbf{Full DSSM-CRF} & \textbf{54.72} & 49.31 & 41.77 & \textbf{66.32} \\
		$-$ frame-level SSM & 51.19 & 46.37 & 38.29 & 64.69 \\
		$-$ dialogue-level SSM & 53.37 & \textbf{50.00} & \textbf{42.27} & 64.92 \\
		$-$ transition residual & 53.83 & 48.78 & 41.57 & 65.71 \\
		$-$ shift supervision & 53.30 & 47.94 & 41.97 & 65.16 \\
		\bottomrule
	\end{tabular*}
\end{table}

\section{Results and Analysis}
\label{sec:results}

\subsection{Overall performance}

Tables~\ref{tab:iemocap_4class_comparison} and~\ref{tab:meld_comparison} compare DSSM-CRF with published systems on IEMOCAP and MELD, respectively. With $\lambda_{\mathrm{shift}}=0.2$, DSSM-CRF achieves 75.81\% UA, 74.90\% WA, and 74.67\% WF1 on the five-fold IEMOCAP evaluation. On MELD, using the validation-selected class-weight exponent $\gamma=0.4$, the model achieves 54.72\% WA and 49.31\% WF1 on the official test set. These results show consistent improvements over the compared audio-based systems across the two datasets despite their different class distributions and conversational structures.

\subsection{Ablation study}

Table~\ref{tab:ablation} evaluates the contribution of individual components and matched structural controls. On IEMOCAP, removing the contextual transition residual causes the largest degradation among the component ablations, reducing UA and WA by 1.72 and 2.11 points, respectively, and lowering Inertia UA by 2.44 points. Removing the dialogue-level SSM or shift supervision also decreases overall performance, while the frame-level SSM has a smaller effect on IEMOCAP.

On MELD, the frame-level SSM contributes most strongly to overall performance: removing it reduces WA and WF1 by 3.53 and 2.94 points, respectively. Removing the dialogue-level SSM, contextual transition residual, or shift supervision also lowers WA. However, some subset and WF1 results do not follow the same trend; in particular, removing the dialogue-level SSM slightly improves WF1 and Shift WA. These results indicate that the proposed components primarily improve overall recognition accuracy while exhibiting different trade-offs across class-weighted and shift-specific metrics. The differing ablation patterns suggest complementary, corpus-dependent roles for fine-grained acoustic modeling and contextual transitions.

\subsection{Shift and inertia behavior}
The full model achieves 59.17\% Shift UA and 79.06\% Inertia UA on IEMOCAP, showing substantially lower recognition accuracy for within-speaker emotion shifts than for persistence. Shift supervision raises Shift and overall UA from 58.19\% and 75.44\% to 59.17\% and 75.81\%, respectively. Removing the contextual transition residual raises Shift UA by 0.57 points but lowers overall and Inertia UA by 1.72 and 2.44 points. Shift performance should therefore be interpreted jointly with overall and Inertia performance.

The matched topology comparison further examines the effect of speaker-wise structured decoding. Without auxiliary shift supervision, the speaker-wise CRF improves UA and WA by 0.99 and 0.91 points over the matched single-chain CRF and improves Inertia UA by 1.49 points, although the single-chain model achieves 1.47 points higher Shift UA. Compared with the retrained no-CRF baseline, the full DSSM-CRF improves UA, WA, and WF1 by 1.17, 1.79, and 2.01 points, respectively. CRF decoding also improves Inertia UA by 1.81 points while reducing Shift UA by 1.86 points, again demonstrating a persistence--shift trade-off. Overall, these matched controls support the benefit of modeling structured emotion transitions along speaker-specific trajectories rather than a single dialogue-wide chain. 

\subsection{Sensitivity and backbone analysis}

\begin{figure}[t]
	\centering
	\includegraphics[width=\columnwidth]{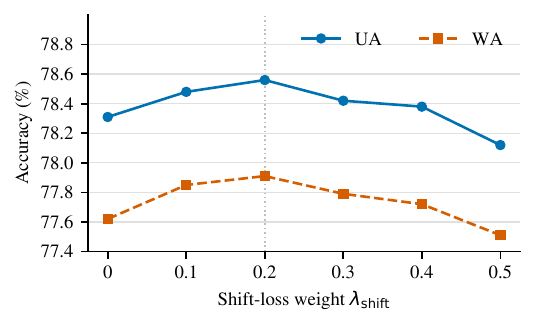}
	\caption{Validation UA and WA across different shift-loss weights on IEMOCAP.}
	\label{fig:shift_sweep}
\end{figure}

Figure~\ref{fig:shift_sweep} shows validation performance across the evaluated shift-loss weights. Both UA and WA peak at $\lambda_{\mathrm{shift}}=0.2$, reaching 78.56\% and 77.91\%, respectively. Across the complete sweep, performance varies by only 0.44 UA and 0.40 WA points, indicating limited sensitivity to the choice of $\lambda_{\mathrm{shift}}$ within the evaluated range.

Table~\ref{tab:backbone} evaluates DSSM-CRF with three self-supervised speech backbones. Relative to the corresponding adapted classifiers, DSSM-CRF improves UA and WA by 15.22 and 18.61 percentage points, respectively, with wav2vec~2.0 Large; by 10.90 and 13.10 points with HuBERT-Large; and by 13.54 and 14.12 points with WavLM-Large. The consistent gains across all three encoders indicate that the proposed conversational modeling framework is not specific to WavLM. The particularly large improvement with wav2vec~2.0 further suggests that dialogue-level contextualization and structured decoding can compensate, at least in part, for weaker utterance-level representations.

\begin{table}[t]
	\centering
	\caption{Backbone robustness on IEMOCAP. Bold indicates the better result for each backbone.}
	\label{tab:backbone}
	\begin{tabular*}{\columnwidth}{@{\extracolsep{\fill}}lccc@{}}
		\toprule
		\textbf{Backbone} & \textbf{Setting} & \textbf{UA(\%)} & \textbf{WA(\%)} \\
		\midrule
		\multirow{2}{*}{wav2vec~2.0 Large}
		& Adapted classifier & 53.83 & 48.27 \\
		& $+$ DSSM-CRF & \textbf{69.05} & \textbf{66.88} \\
		\midrule
		\multirow{2}{*}{HuBERT-Large}
		& Adapted classifier & 60.59 & 57.41 \\
		& $+$ DSSM-CRF & \textbf{71.49} & \textbf{70.51} \\
		\midrule
		\multirow{2}{*}{WavLM-Large}
		& Adapted classifier & 62.27 & 60.78 \\
		& $+$ DSSM-CRF & \textbf{75.81} & \textbf{74.90} \\
		\bottomrule
	\end{tabular*}
\end{table}

\section{Conclusion}
\label{sec:conclusion}

We presented DSSM-CRF, an audio-only framework that separates the structural roles of dialogue-wide contextualization and speaker-specific emotion evolution. Dual-scale bidirectional state-space encoders capture frame-level acoustic dynamics and dialogue-level context, while a speaker-wise dynamic CRF models each speaker's emotion trajectory using a global transition matrix and pair-conditioned residual transitions. Experiments on IEMOCAP and MELD demonstrate strong performance across dyadic and multi-party conversations. Ablations and matched controls support the contributions of contextual transition modeling, speaker-wise chain factorization, and structured CRF decoding, while consistent gains across three self-supervised speech encoders demonstrate backbone robustness. The remaining gap between Shift and Inertia performance highlights emotion transitions as an important direction for further improvement.

\bibliographystyle{template/IEEEbib}
\bibliography{references}

\end{document}